\documentclass{article}

\usepackage{arxiv}
\usepackage{makecell}
\usepackage[utf8]{inputenc} 
\usepackage[T1]{fontenc}    
\usepackage{hyperref}       
\usepackage{url}            
\usepackage{booktabs}       
\usepackage{threeparttable}
\usepackage{amsfonts}       
\usepackage{nicefrac}       
\usepackage{microtype}      
\usepackage{lipsum}		
\usepackage{graphicx}
\usepackage[round]{natbib}
\usepackage{doi}
\usepackage{multirow}
\usepackage{tikz}
\usetikzlibrary{shapes,arrows}

\title{\textsc{DirectQuote:} A Dataset for Direct Quotation Extraction and Attribution in News Articles}

\author{ {Yuanchi Zhang$^1$ and Yang Liu$^{1,2,3,4,5}$} \\
	$^1$Department of Computer Science and Technology, Tsinghua University, Beijing, China\\
	$^2$Institute for AI Industry Research, Tsinghua University, Beijing, China\\
$^3$Institute for Artificial Intelligence, Tsinghua University, Beijing, China \\
	$^4$Beijing National Research Center for Information Science and Technology \\
	$^5$Beijing Academy of Artificial Intelligence \\
}

\renewcommand{\shorttitle}{DirectQuote: A Dataset for Direct Quotation Extraction and Attribution in News Articles}

\hypersetup{
pdftitle={DirectQuote: A Dataset for Direct Quotation Extraction and Attribution in News Articles},
pdfauthor={Yuanchi Zhang, Yang Liu},
pdfkeywords={Quotation Extraction, Quotation Attribution, News Corpora},
}

\begin{document}
\maketitle

\begin{abstract}
Quotation extraction and attribution are challenging tasks, aiming at determining the spans containing quotations and attributing each quotation to the original speaker. Applying this task to news data is highly related to fact-checking, media monitoring and news tracking. Direct quotations are more traceable and informative, and therefore of great significance among different types of quotations. Therefore, this paper introduces \emph{DirectQuote}, a corpus containing 19,760 paragraphs and 10,279 direct quotations manually annotated from online news media. To the best of our knowledge, this is the largest and most complete corpus that focuses on direct quotations in news texts. We ensure that each speaker in the annotation can be linked to a specific named entity on Wikidata, benefiting various downstream tasks. In addition, for the first time, we propose several sequence labeling models as baseline methods to extract and attribute quotations simultaneously in an end-to-end manner.\footnote{The DirectQuote corpus is available at  \url{https://github.com/THUNLP-MT/DirectQuote}}
\end{abstract}


\section{Introduction}
\begin{figure}[t]
\centering     \includegraphics[width=16cm]{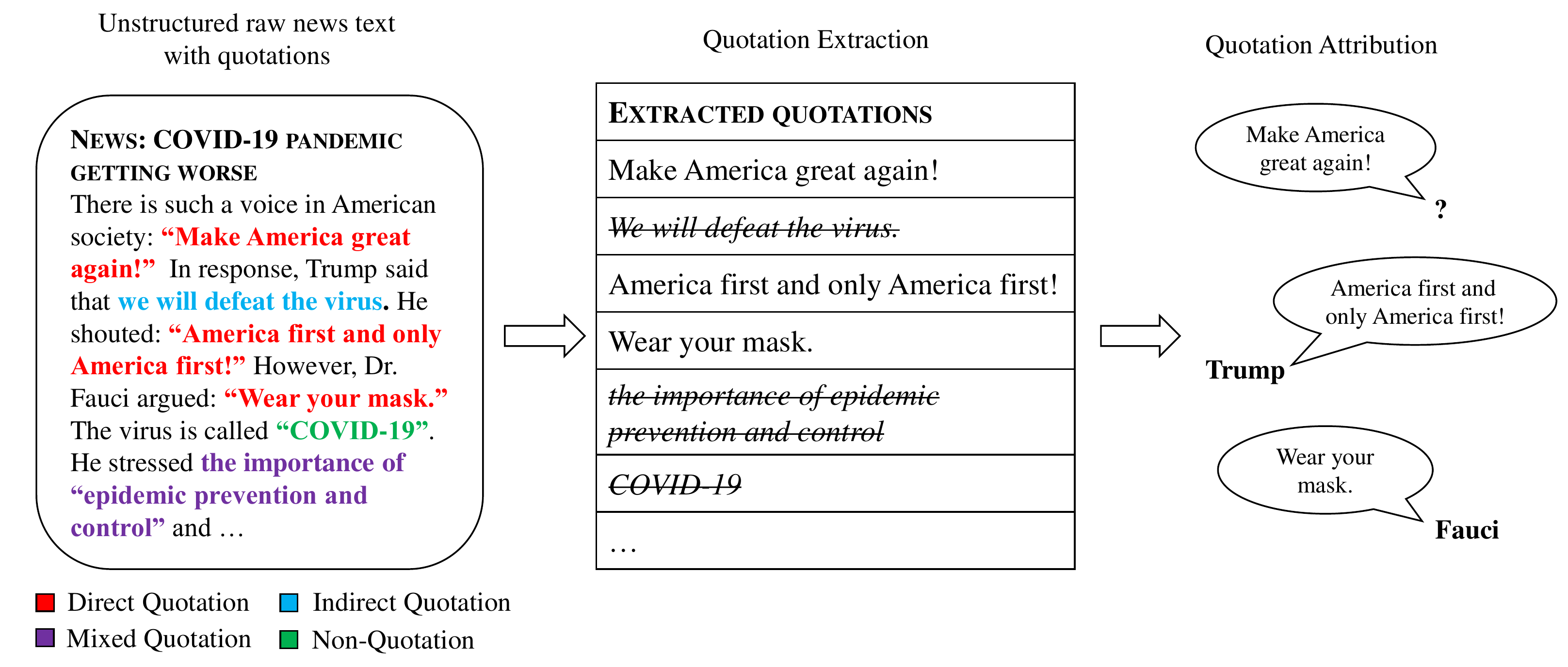}     \caption{Direct quotation extraction and attribution in news articles.}     \label{fig:overview} \end{figure}

A \emph{quotation} is a general notion that covers different kinds of speech, thought, and writing in text \citep{semino2004corpus}. It is a prominent linguistic device for expressing opinions, statements, and assessments attributed to the speaker \citep{cappelen2012quotation}. Concretely, there are three types of quotations: \emph{direct quotations}, \emph{indirect quotations} and \emph{mixed quotations} (Table \ref{fig:overview}). Among all kinds of quotations, the entire content of the \emph{direct quotation} \citep{o2013annotated} is in quotation marks, which means that what the speaker said is transcribed verbatim. 

Direct quotations are of considerable sigificance among all types of quotations. On the one hand, news writers attribute direct quotations to speakers, making claims credible and authoritative, leading to more traceable and informative news \citep{wp2020understanding}. These direct quotations from politicians, public figures, and other celebrities improve the authenticity and fairness of news, making news more convincing. On the other hand, the development of social media and advanced language generation models, such as GPT, has led to the proliferation of fake news \citep{floridi2020gpt}. Troubled by rampant media manipulators, people are increasingly questioning the political stance and legitimacy of the news press. Therefore, direct quotations in the news are essential to ensure the transparency and accountability of news.

Besides news writing, direct quotations are also actively  involved in various NLP tasks. Since direct quotations are high subjective opinions, cognitions and assertions, they are used in opinion mining and claim detection task \citep{balahur2009opinion} to discover opinions, sentiment analysis task \citep{balahur2013sentiment} to evaluate the author’s mood, fact check task to verify factual information, statement monitoring task to track others’ speech. Many websites apply these tasks. For example, NewsBrief is a website that automatically extracts and attributes quotations, detects events, and updates them in real-time. Another website, Politifact, tracks the statements of politicians for news fact-checking to reduce misinformation. ISideWith tracks political views on different topics to boost voter engagement and education. However, these systems rely heavily on costly and time-consuming human labor.

In general, as shown in Figure \ref{fig:overview}, the above applications include two types of tasks related to quotations. The first task is called \emph{quotation extraction} that refers to determining the span that represents the quotation in the document. The second task, \emph{quotation attribution}, refers to determining the speaker of the quotation. It is one of the primary criteria for maintaining integrity in journalism as a primary rhetorical mechanism to promote the veracity and correctness of reporting.

However, insufficient attention has been paid to direct quotations. Note that several research focus on various quotations. Most of the related corpora \citep{papay2020riqua,stymne2020slanda,chen2019chinese,zhang2019whose} are based on novels. We find that literary works such as novels usually have fixed speakers and are sophisticated grammatically and semantically. In contrast, corpora such as PARC\citep{pareti2016parc} and Polnear\citep{newell2018attribution} are annotated in news text, which follows a more straightforward grammatical style and has varying speakers. While these pioneering work has greatly facilitated the research on the quotation, the task of quotation extraction and attribution still faces many challenges.

On the one hand, large-scale deep learning models, such as BERT and GPT, severely overfit on these corpora with limited size. Furthermore, corpora generated by rule-based methods lack diversity, limiting the generalization ability of the model. On the other hand, some corpora study broader attribution relations that take propositional attitudes including quoting and paraphrasing into consideration. This definition goes beyond the scope of direct quotations, leading to inconsistencies with downstream tasks, such as fact check and journalism supervision. Moreover, some annotated speakers are ambiguous and adversely affect downstream tasks such as entity linking.

To alleviate the above problems, we introduce a corpus of direct quotations called \emph{DirectQuote}. To build the corpus, we continuously sample news from multiple news sources to keep the text distribution in the corpus consistent with that in actual applications. Based on the data, we select 19,706 paragraphs containing quotation marks, and annotate 10,279 quotations and corresponding speakers. When annotating speakers, we ensure that valid speakers should be linked to a person entity in a named entity library. Among them, simple patterns are removed to increase the diversity of the corpus. To the best of our knowledge, it is the largest direct quotation extraction and attribution corpus. We hope that the corpus can assist people in understanding and analyze quotations in the news.

\section{Related Work}
Quotations usually contain propositional attitudes toward a third party, which is a common form of attribution. The statements, intentions, beliefs, knowledge, cognition, and decrees contained in the quotations belong to the scope of event detection, discourse relations, and opinion analysis.

Many comprehensive corpora contain annotations related to quotations and attributions; however, these corpora do not directly annotate quotations, so quotations and attributions are studied as part of other tasks and are limited to specific categories, resulting in a low recall. Sometimes,  the speaker of the quotation is not explicitly marked. For example, the rhetorical annotations in PDTB \citep{pdtb} potentially contain quotation attributions; TimeBank \citep{pustejovsky2003timebank} only focuses on attributions related to events; MPQA \citep{mpqa} and NTCIR \citep{ntcir7,ntcir8} extract quotations containing opinions, opinions, and emotions; NTCIR is composed of quotations with explicit sources; CQSA \citep{cqsa} extracted the direct quotations in the novel. PolNEAR is a corpus of news articles in English built by  \cite{newell2018attribution}. Built on a corpus sampled from the coverage of an event of great sociological, political, and journalistic import-campaign coverage, this corpus is an attribution corpus that focuses on attribution relations in a very broad sense.

Some corpora are dedicated to quotation extraction and attribution.  \cite{elson2010extracting} assemble a corpus of more than 3000 quotations from a collection of works by six modern authors and manually label the speakers;  \cite{o2013annotated} annotate 3,535 direct quotations and sources from 965 newswire texts of Sydney Morning Herald. PARC is a corpus annotated with attribution relations based on Wall Street Journals and can be used to analyze attribution and validate assumptions.  \cite{fernandes2011quotation} annotate the quotations in 685 articles based on GloboQuotes, an unlabeled Portuguese corpus obtained from the Global.com website. Using IOB coding, the corpus is codified in a per token basis and built with golden annotation for named entities, coreferences, quotations, and associations between quotations and authors; however, the scale of English news quotations in these corpora is limited.

\section{Corpus Construction}
Figure \ref{fig:flow} shows the flowchart of constructing the corpus. In this section, we will introduce the main steps to construct the corpus in detail.

\subsection{Collection}

\begin{table}
	\caption{Data in the \emph{DirectQuote}.}
	\centering
\begin{tabular}{lllr}
\toprule
Region      & Name & URL &Numbers                               \\
\midrule
\multirow{4}{*}{U.S.}      & Associated Press & \url{https://apnews.com} & 438\\
& Cable News Network& \url{https://edition.cnn.com} &627\\
& American Broadcasting Company& \url{https://abcnews.go.com} &240 \\
& New York Times& \url{https://www.nytimes.com} &5,642 \\
& CBS Broadcasting& \url{https://www.cbsnews.com} &4,890\\
\midrule
\multirow{3}{*}{UK}            & British Broadcasting Corporation   & \url{https://www.bbc.com}  &926      \\
&Reuters& \url{https://www.reuters.com} &5,836 \\
&The Guardian& \url{https://www.theguardian.com} & 4,302\\
\midrule
\multirow{2}{*}{Canada}        & The Globe and Mail     & \url{https://www.theglobeandmail.com}     &1,955             \\
& The Star & \url{https://www.thestar.com.my} & 13,769\\
\midrule
New Zealand & NZ Herald  & \url{https://www.nzherald.co.nz}  &115                         \\
\midrule
\multirow{2}{*}{Australia}     & Australian Broadcasting Corporation & \url{https://www.abc.net.au}   & 312                    \\
& Sydney Morning Herald& \url{https://www.smh.com.au} & 93\\
\bottomrule
\end{tabular}
	\label{tab:source_table}
\end{table}

We hope that the quotations in the corpus are diverse and can be applied in various downstream tasks such as stance analysis and sentiment analysis. Therefore, we select representative and multiple news sources across the political spectrum, including 13 well-known online news media from five major English-speaking countries, as shown in Table \ref{tab:source_table}.

We start by collecting publicly available news from each publisher between September 2020 and March 2021 from the Internet. After deduplication and removal of unqualified news, we obtain 40,000 news articles on different topics and genres. These texts are all downloaded from the publisher's official website using a web crawler. When obtaining news, we collect the necessary metadata to ensure traceability: news title, publication time, publisher name, and source URL.

\subsection{Processing}
\begin{figure}[t]
\centering     \includegraphics[width=16cm]{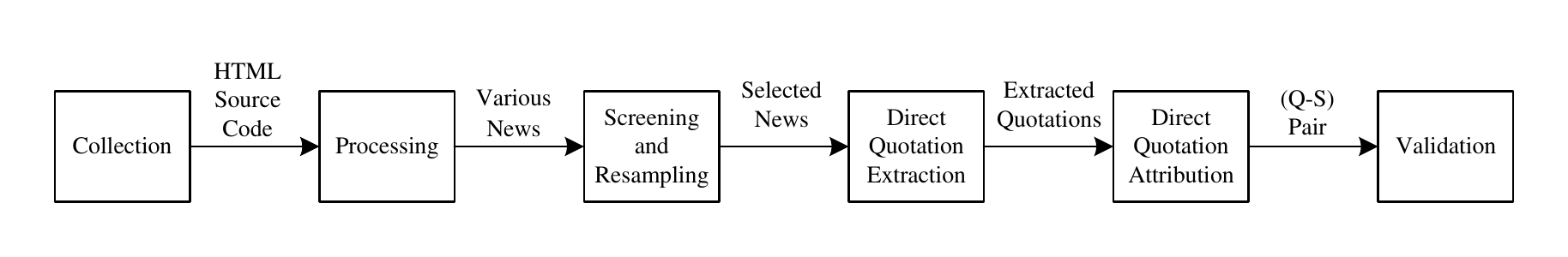}     \caption{Flowchart of corpus construction. ``Q-S Pair'' stands for ``Quotation-Speaker Pair''.}     \label{fig:flow} \end{figure}

Considering that pre-training models such as BERT usually use fixed-size vocabulary, we need to ensure that the input text contains only common English letters and punctuations to  facilitate training and fine-tuning. Therefore, we extracted the plain text and related metadata from the source code of the webpage downloaded from a website that is mixed with a large number of HTML tags.

A single line break ("$\backslash$n") is used in the text to distinguish different paragraphs. Rich media interspersed in the original webpage, such as photos, video clips, audio clips, tweets, and other web pages, are deleted. The advertisements in the article and the copyright notice placed at the end of the article have also been removed so as not to affect the model's understanding of semantics. The CSS styles used by some web pages to add punctuation marks, such as quotes to news, are converted into standard characters. Invalid characters, such as Emoji and non-English characters, are replaced with spaces. The title of each subsection is kept in a separate paragraph.

\subsection{Screening and Resampling}
The news collected covers many genres. Hard news focuses on significant current events, while soft news focuses on editorials, blogs, summaries, and gossip. Serious news covers political and business news, while entertainment news is about sports, fashion, and travel. These texts differ in style, semantics, and grammar. The length of the article, the number of entities, and the number of quotations vary greatly, so the difficulty of quotation extraction and attribution tasks is different.

To alleviate the unbalanced distribution of news text in time, length, genre, and content, we filter and resample the collected news. After these processes, we obtain hard news covering global politics that is evenly distributed in terms of publication time and the number of words; moreover, publishers such as the New York Times tend to combine reports and opinions in the same text. These texts are also selected into the corpus to increase the diversity of the corpus and prevent the model from deteriorating when dealing with such news.

\subsection{Direct Quotation Extraction}

\emph{Quotation extraction} is defined as extracting reported speech from a third party in the text, also known as reported speech extraction. As shown in Table \ref{fig:flow}, there are three types of quotation based on the position of the quotation marks. The entire content of the \emph{direct quotation} is in quotation marks, which means that what the speaker said is transcribed verbatim. In contrast, quotation marks do not identify \emph{indirect quotations} that change the speaker's original words but have the same meaning. In particular, part of the content of a \emph{mixed quotation} is inside the quotation marks, and part is outside the quotation marks, which means that the critical content of the speaker is kept verbatim, and the rest is adjusted to fit the sentence grammatically. 

The quotation marks have other usages. Some writers put quotation marks around the words they want to distance themselves from. Quotation marks may also indicate words used ironically or with some reservations or to signify words used as words. In these cases, the text inside the quotation marks is not a quotation.

First, we take direct and mixed quotations into consideration; however, mixed quotations are often confused with other sentence components, which seriously affects the token level agreement. The distinction between mixed speech and unquoted speech is also vague.

To simplify the extraction task, we only consider the direct quotations containing independent sentences, which means that there is at least one clause in the quotation marks. This provision ensures that the extracted text expresses something, which is conducive to applying quotations in downstream tasks such as opinion mining and sentiment analysis. In the implementation, simple regular expressions are used to extract all of the text inside the quotations. Thereafter, the quotations that meet the requirements are selected through manual annotation.

\subsection{Direct Quotation Attribution}

\renewcommand{\arraystretch}{1.5}
\begin{table}
	\caption{Representative examples of direct quotation extraction and attribution}
	\centering
\begin{tabular}{|p{15.5cm}|}
\hline
 President \textbf{Trump} who made a statement that \emph{“the growth was about to peak”} has been visiting China. \\
\hline
Many said: \emph{“Release the prisoners!”} \\
\hline
Rothstein said she rejected “any suggestion that the public interest favors requiring AWS to host the incendiary speech that the record shows some of Parler’s users have engaged in.” \\
\hline
\emph{“If you find it in Germany,”} \textbf{he} said, \emph{“you will find it in my bakery.”} \\
\hline
\end{tabular}
	\label{tab:case}
\end{table}

In this step, each quotation is attributed to an optional speaker. Specifically, the speaker is determined by a text span in the context window where the quotation is located.

Because the quotation attribution task mainly tracks people's attitudes and statements, a large number of downstream tasks, such as  fact-checking,  will perform entity linking on speakers. Therefore, we hope that the speaker is a mention of an exact person entity. Hence, every speaker can be linked to an item in an entity library, such as Wikidata. Specifically, the legal speaker span can be a personal pronoun or a person's name. Allowing pronouns to act as legitimate speakers separates the coreference resolution from the attribution task, which improves the token-level agreement and reduces the possibility that the speaker cannot be found in a smaller context window. Therefore, the task is simplified,  and the sequence length is shortened.

The speaker is neither a broad communicative agent, such as an artifact, organization, website, database, or the implication of a person, such as identity, nickname, and characteristics, because what they refer to may not be associated with a specific person. Low traceability hinders the accuracy of the entity-linker.

Speakers' spans are manually annotated. The annotators from the crowdsourcing platform determine an optional speaker for each quotation in the context window. If the annotator fails to find a specific speaker (no speaker or no qualified speaker) in the context window, the quotation will be assigned the \emph{NONE} label. If multiple legal speakers lexically indicate the attribution relation, the one closest to the quotation is selected. The quotation span and  speaker span must be continuous. Multiple quotations share the speaker's span. The annotation format is a text pair (quote, speaker). This task ignores nested quotations.

We show some examples to demonstrate the direct quotation extraction and attribution (Table \ref{tab:case}). The first example is a typical pattern of direct quotation extraction and attribution. We choose ``Trump'' instead of ``President Trump'' as the speaker, because ``President'' is an identity, not part of his name. Hence we get the most accurate speaker span. In the second example, ``many'' is not a named entity or unambiguous pronoun, therefore it is not a valid speaker. For this reason, the quotation does not has a speaker. By adding the requirement, we ensure that all speakers are traceable and suitable for downstream tasks such as entity linking. In the third example, content in quotation marks is a none phrase instead of a complete clause. 
The premise that quotations annotated have clear semantics is guaranteed by excluding incomplete sentence. In the fourth example, two quotations share the same speakers. The sequence labeling method described above is capable of handling this pattern.

\subsection{Validation}
During annotation, we revise the annotation guidelines to improve the consistency of the corpus. At least three crowdsourced workers view the labels of each context window. If an annotation decision cannot be agreed upon, it will be determined according to the annotation guidelines. For context windows with multiple candidate speakers, the author manually check all annotations. Finally, in 19,706 context windows, 10,279 quotations are obtained, of which 8,831 can be attributed to a specific speaker, and the remaining 1,522 have no explicit valid speakers.

We treat quotation extraction and attribution as a sequence labeling task and treat each token as an independent labeling decision to examine whether the label is correct and whether the boundary of each span is accurate. After sampling 20$\%$ of the context windows, the Krippendorff's alpha is 0.81. The vast majority of inconsistencies are due to the ambiguity of the speaker span.

\begin{figure}     \centering     \includegraphics[width=16cm]{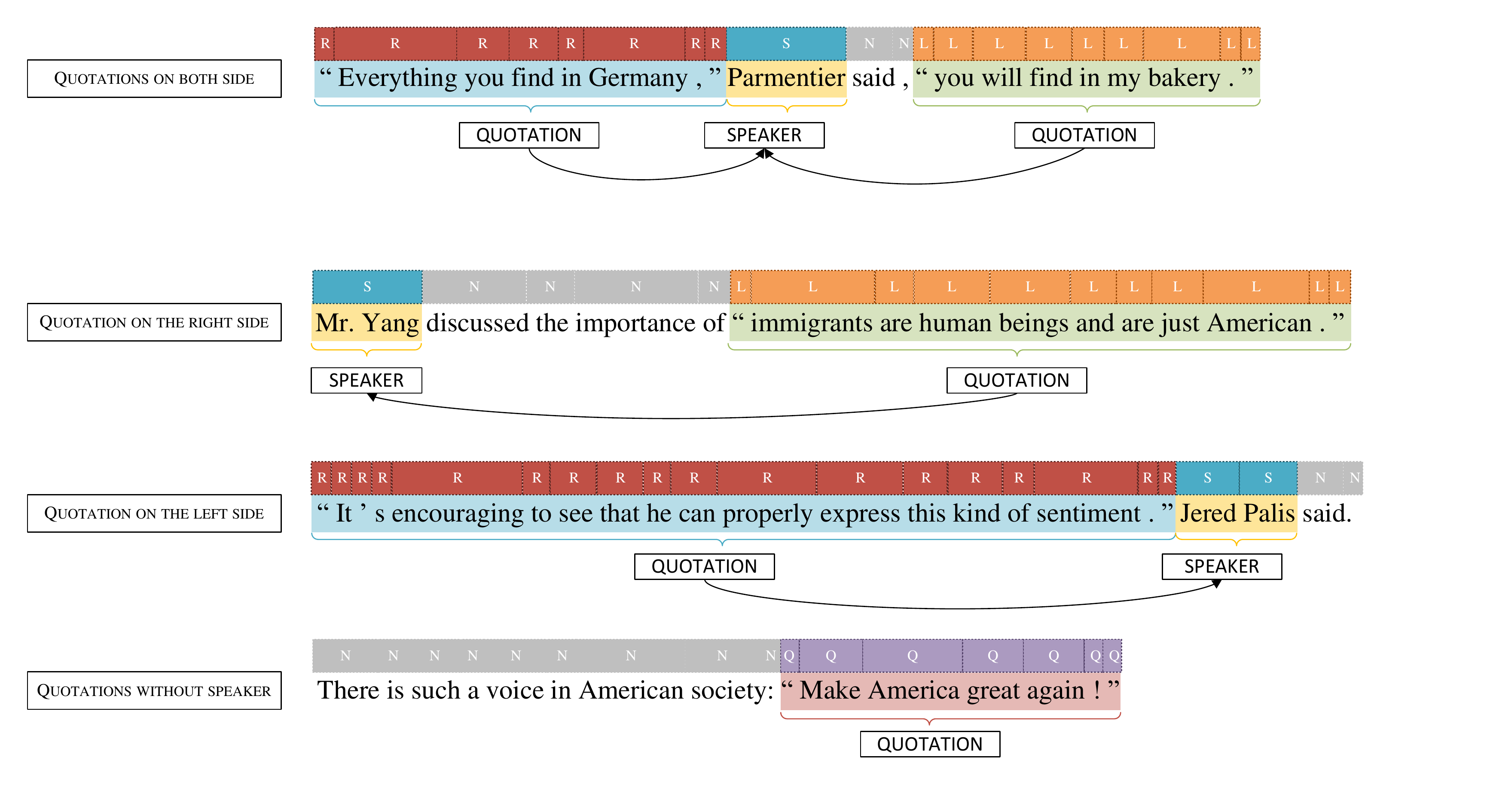}     \caption{Relative position of the quotation and the speaker and the corresponding sequence.}     \label{fig:color} \end{figure}

\section{Baseline Methods and Results}

\subsection{Methods}

Although the common rule-based method is fast and performs well in the cases of simple syntax, it cannot handle rare quotation patterns.

We decide not to adopt the pipeline design of \cite{pareti2016parc}. In their work, separate models are designed to solve quotation extraction and attribution tasks, and additional models are introduced for named entity recognition and dependency analysis during preprocessing. Multiple pipeline stages increase the cumulative error and increase the time and space cost of the model, which is not suitable for scenarios with massive amounts of data.

Instead, an end-to-end sequence annotation model is designed to perform quotation extraction and attribution tasks simultaneously. Our model takes the original text as input and directly predicts the quotation and speaker.  As shown in Figure \ref{fig:color}, the model not only outputs the span of the quotation and  speaker, but also determines the correspondence between the quotation and speaker by predicting the speaker's direction relative to the quotation.

As shown in Figure \ref{fig:color}, quotation extraction and attribution tasks are framed as sequence labeling tasks, and each token is classified into one of the following labels:
\begin{itemize}
    \item  Quotation, the corresponding speaker is in the preceding text (denoted by ``\includegraphics[scale=0.4]{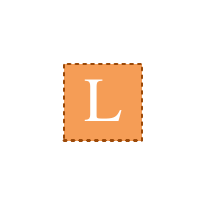}'')
    \item  Quotation, the corresponding speaker is in the following text (denoted by ``\includegraphics[scale=0.4]{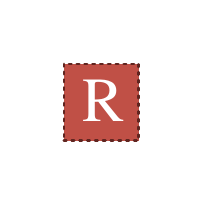}'')
    \item  Quotation, no corresponding speaker (denoted by  ``\includegraphics[scale=0.4]{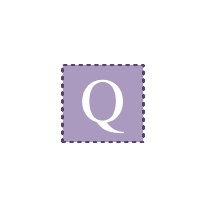}'')
    \item Speaker (denoted by  ``\includegraphics[scale=0.4]{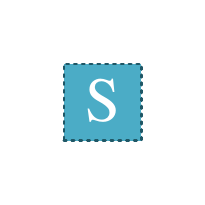}'')
    \item Neither (denoted by  ``\includegraphics[scale=0.4]{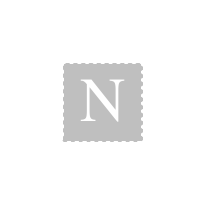}'')
\end{itemize}
All labels are in the IOB1 format. According to the type of quotation label (left, right, no speaker), we can determine the attribution of the quotation. In the corpus, the proportions of three types of quotations, ``\includegraphics[scale=0.4]{l.pdf}''``\includegraphics[scale=0.4]{r.pdf}''``\includegraphics[scale=0.4]{q.pdf}'' , are 46\%, 39\% and 15\% respectively.

We use the following sequence labeling methods and components:
\begin{itemize}

\item \textbf{CRF}  First proposed by \cite{lafferty2001conditional}, conditional random field (CRF) is a type of probabilistic graphical model to model sequential data such as labels and words in sentences. During training, CRF will determine the weights of hand-crafted feature functions to predict the labels. We adopt CRF as a type of non-neural network method.

\item \textbf{LSTM} Long Short-Term Memory (LSTM) is a redesign of traditional recurrent neural network architecture around its memory cell \citep{hochreiter1997long}. LSTM has been shown capable of storing and accessing information over very long timespans in varied sequence labeling tasks such as POS tagging. We train a LSTM model as a non-pretrained neural network model.

\item \textbf{CNN} Convolutional neural network is successful in some NLP tasks using convolution and max-pooling layer to summarize inputs \citep{ma2016end}. CNN runs in parallel to extract  extracting local spatial and temporal dependence features. Therefore, it is faster than LSTM. We add CNN layer to our model to further enhance ability in feature extraction.

\item \textbf{BERT} \cite{devlin2018bert} develop BERT, a pre-trained transformer-based language model that achieves excellent performance through simple fine-tuning on various natural language understanding tasks. 
Compared with previous methods, BERT has stronger generalization ability, which can effectively extract semantic and grammatical features considering long-distance dependencies, but also involves massive parameters that occupy a lot of storage space and take a long time to run once.
\end{itemize}

All model weights are fine-tuned to maximize the log-likelihood of the output corresponding to the ground-truth label using the cross-entropy loss. Neither BERT nor LSTM can handle text that is too long, so each paragraph in the entire text is determined as a context window to be input into the model. 

To prevent the model from overfitting  frequently occurring names, we randomly replace speakers with other names for data augmentation.

\subsection{Results}
\renewcommand{\arraystretch}{1}
\begin{table}
  \caption{Results of quotation extraction and attribution, calculated in three categories.}
  \centering
  \fontsize{8}{10}\selectfont
  \begin{threeparttable}
  \label{tab:performance_comparison}
    \begin{tabular}{lccccccccc}
    \toprule
    \multirow{2}{*}{Method}&
    \multicolumn{3}{c}{ Speaker}&\multicolumn{3}{c}{ Quotation}&\multicolumn{3}{c}{ Overall}\cr
    \cmidrule(lr){2-4} \cmidrule(lr){5-7} \cmidrule(lr){8-10}
    &Precision&Recall&F1-Measure&Precision&Recall&F1-Measure&Precision&Recall&F1-Measure\cr
    \midrule
    CRF&0.7307&0.5296&0.6141&0.6969&0.6918&0.6943&0.6977&0.6864&0.6920\cr    LSTM&0.7368&0.8286&0.7801&0.7711&0.8534&0.8102&0.7700&0.8525&0.8092\cr
    LSTM+CRF&0.7282&0.7931&0.7593&0.7816&0.8957&0.8348&0.7799&0.8922&0.8323\cr   
    LSTM+CNN&0.7150&0.8598&0.7808&0.8188&0.9135&0.8635&0.8150&0.9117&0.8606\cr
    LSTM+CNN+CRF&0.7492&0.8409&0.7924&0.8129&0.9263&0.8659&0.8107&0.9234&0.8634\cr
    Bert-Base&{\bf 0.8090}&{\bf 0.9402}&{\bf 0.8697}&{\bf 0.8169}&{\bf 0.9354}&{\bf 0.8721}&{\bf 0.8164}&{\bf 0.9356}&{\bf 0.8720}\cr
    \bottomrule
    \end{tabular}
    \end{threeparttable}
\end{table}
The precision, recall, and F1 scores of the baseline method are shown in Table \ref{tab:performance_comparison}. 70$\%$ of the corpus is divided into training set, and the remaining 30$\%$ is divided into test set. The results are shown in three categories: speaker, quotation, and overall. Almost all models have F1 scores of 70$\%$ or higher in the quotation extraction and attribution tasks. This proves that it is feasible to model the joint task of quotation extraction and attribution as a sequence labeling task. By adding a CNN layer or a CRF layer, the feature extraction ability of LSTM is improved, which increases the overall recall rate by approximately 3$\%$-6$\%$. Because BERT is pre-trained on a large number of corpora, and the Transformer model can effectively extract different levels of semantic and grammatical features, the BERT model has stronger accuracy and generalization capabilities, and outperforms other models by approximately 6$\%$ absolute precision and recall of the speaker.

\subsection{Case Study}
\renewcommand{\arraystretch}{1.2}
\begin{table}[t]
	\centering
	\caption{Examples of extraction and attribution results of the baseline models. Quotations are marked in blue, and speakers are marked in red.}

\begin{tabular}{l|p{14cm}}
\hline \hline
\textbf{Input} & “We are concerned the failure to secure an adequate supply of vaccines will needlessly prolong the COVID-19 pandemic in this country, causing further loss of life and economic devastation,” a group of senators led by Patty Murray of Washington and Ron Wyden of Oregon wrote HHS.  \\
\hline 
\textbf{Ground Truth} & \textbf{\textcolor{cyan}{“We are concerned the failure to secure an adequate supply of vaccines will needlessly prolong the COVID-19 pandemic in this country, causing further loss of life and economic devastation,”}} a group of senators led by Patty Murray of Washington and Ron Wyden of Oregon wrote HHS.  \\
\hline
\textbf{CRF} & \textbf{\textcolor{cyan}{“We are concerned the failure to secure an adequate supply of vaccines will needlessly prolong the COVID-19 pandemic in this country, causing further loss of life and economic devastation,”}} a group of senators led by Patty Murray of Washington and \textbf{\textcolor{red}{Ron Wyden}} of Oregon wrote HHS. \\
\hline
\textbf{LSTM} & \textbf{\textcolor{cyan}{“We are concerned the failure to secure an adequate supply of vaccines will needlessly prolong the COVID-19 pandemic in this country, causing further loss of life and economic devastation,”}} a group of senators led by \textbf{\textcolor{red}{Patty Murray}} of Washington and \textbf{\textcolor{red}{Ron Wyden}} of Oregon wrote HHS.
\\
\hline
\textbf{BERT} & \textbf{\textcolor{cyan}{“We are concerned the failure to secure an adequate supply of vaccines will needlessly prolong the COVID-19 pandemic in this country, causing further loss of life and economic devastation,”}} a group of senators led by Patty Murray of Washington and Ron Wyden of Oregon wrote HHS.
\\
\hline \hline
\textbf{Input} & DHS said FEMA will fund 31 high-threat, high-density urban areas that will ``be required to dedicate a minimum of 30\% of awards toward five priority areas: cybersecurity (7.5\%); soft target and crowded places (5\%); information and intelligence sharing (5\%); domestic violent extremism (7.5\%); and emerging threats (5\%).'' \\
\hline 
\textbf{Ground Truth} & DHS said FEMA will fund 31 high-threat, high-density urban areas that will ``be required to dedicate a minimum of 30\% of awards toward five priority areas: cybersecurity (7.5\%); soft target and crowded places (5\%); information and intelligence sharing (5\%); domestic violent extremism (7.5\%); and emerging threats (5\%).''  \\
\hline
\textbf{CRF} & \textbf{\textcolor{red}{DHS}} said FEMA will fund 31 high-threat, high-density urban areas that will \textbf{\textcolor{cyan}{``be required to dedicate a minimum of 30\% of awards toward five priority areas: cybersecurity (7.5\%); soft target and crowded places (5\%); information and intelligence sharing (5\%); domestic violent extremism (7.5\%); and emerging threats (5\%).''}} \\
\hline \textbf{LSTM} & \textbf{\textcolor{red}{DHS}} said FEMA will fund 31 high-threat, high-density urban areas that will \textbf{\textcolor{cyan}{``be required to dedicate a minimum of 30\% of awards toward five priority areas: cybersecurity (7.5\%); soft target and crowded places (5\%); information and intelligence sharing (5\%); domestic violent extremism (7.5\%); and emerging threats (5\%).''}} \\
\hline
\textbf{BERT} & DHS said FEMA will fund 31 high-threat, high-density urban areas that will ``be required to dedicate a minimum of 30\% of awards toward five priority areas: cybersecurity (7.5\%); soft target and crowded places (5\%); information and intelligence sharing (5\%); domestic violent extremism (7.5\%); and emerging threats (5\%).'' \\
\hline \hline
\end{tabular}
	\label{tab:case_study}
\end{table}

Two examples are shown in Table \ref{tab:case_study}. In the first instance, CRF and LSTM fail in judging the speakers. It is quite a confusing case because ``Patty Murray'' and ``Ron Wyden'' are actually leaders of the speakers instead of speakers themselves. Note that BERT is more capable of understanding semantics while other models simply extract all or part of named entities. The second example illustrates that a quotation must not be an incomplete sentence in quotation marks. CRF and LSTM-based models, which misjudge the content as a quotation, may extract quotations based mainly on quotation marks. The output of BERT is correct here probably because BERT is more sensitive to syntax.

\section{Conclusion}
This paper presents a new corpus, \emph{DirectQuote}, where news from 13 publishers in five countries is manually annotated to extract direct quotations and corresponding speakers. It is helpful in solving the limitation of insufficient resources in the quotation extraction and attribution tasks and enable researchers to propose more sophisticated models.
We apply multiple sequence labeling models to the joint task of quotation extraction and attribution as end-to-end baseline methods. The pre-trained BERT model performs better in terms of accuracy and recall.
By increasing the scale of the dataset, extracting mixed quotations and indirect quotations, and applying the dataset to large-scale automatic extraction and attribution systems will be the future research direction.

\section*{Acknowledgement}
The authors would like to thank Xuancheng Huang and Fanzeng Xia for reviewing and giving feedback on drafts of the paper.

\bibliographystyle{unsrtnat}
\bibliography{myref.bib}

\end{document}